\documentclass[journal]{IEEEtran} 
\IEEEoverridecommandlockouts

\usepackage{cite}

\usepackage[pdftex]{graphicx}
\usepackage{subcaption}

\usepackage{xcolor}

\def\BibTeX{{\rm B\kern-.05em{\sc i\kern-.025em b}\kern-.08em
    T\kern-.1667em\lower.7ex\hbox{E}\kern-.125emX}}

\begin{document}

\title{Data Amount Heterogeneity Improves Performance in Federated Learning for Neuroimaging}

\author{
    \IEEEauthorblockN{Paul Kurian\textsuperscript{1}}, 
    \IEEEauthorblockN{Tamoghna Chattopadhyay\textsuperscript{2}},
    \IEEEauthorblockN{Abhijith Shaji\textsuperscript{1}},\\
    \IEEEauthorblockN{Greg Ver Steeg\textsuperscript{3}},
    \IEEEauthorblockN{Paul M. Thompson\textsuperscript{2}},
    \IEEEauthorblockN{Jos\'{e} Luis Ambite\textsuperscript{1}}
    \\
    \IEEEauthorblockA{\textsuperscript{1}\textit{Information Sciences Institute, University of Southern California, Marina del Rey, CA, USA}}\\
    \IEEEauthorblockA{\textsuperscript{2}\textit{Imaging Genetics Center, Mark and Mary Stevens Neuroimaging and Informatics Institute, Keck School of Medicine, University of Southern California, Marina del Rey, CA, USA}}\\
    \IEEEauthorblockA{\textsuperscript{3}\textit{University of California, Riverside, CA, USA}}\\
    \IEEEauthorblockA{ambite@isi.edu}
}

\maketitle

\begin{abstract}
 Federated Learning (FL) allows for secure analysis of distributed data \textit{without data sharing}. We and others have successfully applied FL to neuroimaging tasks, such as Alzheimer's disease classification from structural magnetic resonance imaging (MRI). FL trains a deep learning model locally at each site, encrypts the local model parameters and sends them to a federation controller that aggregates the local models under homomorphic encryption, the resulting community model is sent back to the sites for further training, and the process repeats for a number of rounds. FL often achieves learning performance (e.g., accuracy) close to centralized training, but no data leaves the sites and model parameters are protected through encryption and other measures. 

 In this paper, we show that federations composed of sites with heterogeneous amounts of data (1 large site plus small sites) perform better than homogeneous (same data per site) federations, particularly in low data regimes, while reaching performance close to centralized training. Having a big site often results in faster convergence and higher final accuracy. 
 Heterogeneous federations better exploit sites with small amounts of data, a common situation in large consortia such as ENIGMA.

\end{abstract}

\section{Introduction}

Dementia currently affects approximately 55 million people worldwide, with that number projected to reach 139 million by 2050. Alzheimer's disease (AD) accounts for 60–80\% of all cases and has become the seventh leading cause of death globally [1]. Despite this scale, reliable automated tools for distinguishing individuals with dementia from cognitively normal (CN) controls remain scarce in clinical deployment, largely because training robust deep learning models requires access to large, diverse neuroimaging datasets that span multiple acquisition sites, scanner types, and patient populations. 

The inherent complexity and cost of enforcing security and privacy results in few large-scale data sharing efforts. Even when large consortia are established, they often only perform meta-analysis using traditional statistical methods. A paradigmatic example of large-scale meta-analysis is the ENIGMA Consortium~\cite{thompson2020enigma}. 
In contrast, Federated Learning enables secure, distributed \textit{mega-analysis} using deep learning methods.

\section{Federated Learning}

Federated Learning~(FL)~\cite{mcmahan2017communication,yang2019federated,li2020federated,jordan2018communication,rieke2020future} enables large-scale cross-institutional data analysis while maintaining the data private at each site. 
Federated Learning is being increasingly applied in biomedical  domains~\cite{lee2018privacy,sheller2018multi,silva2019federated,rieke2020future,silva2020fed}.
Similarly, much recent work has used Federated Learning for distributed neuroimaging data analysis~\cite{li2020multi,rootes2022federated,stripelis2021secure,fan2021federated,stripelis2024:patterns,mitrovska2024secure}.

Federated Learning was introduced by McMahan et al. \cite{mcmahan2017communication} for learning over user data in mobile phones. Their original algorithm, \textit{Federated Average (FedAvg)}, follows a synchronous communication protocol where each learner (phone) trains a neural network for a fixed number of epochs on its local dataset. Once all learners (or a subset) finish their locally assigned training, the federation controller computes a community model that is a weighted average of each of the learners' local models, with the weight of each learner proportional to the number of its local training examples. The new community model is then distributed to all learners and the process repeats. 

We recently developed a federated training policy, \textit{SemiSync}~\cite{stripelis2022:semisync}, which is particularly useful in environments with heterogeneous data distributions and computational capabilities. In SemiSync, the learners' aggregation point is not based on a fixed number of epochs, but on a synchronization period based on the processing speed of the slowest site. Learners with different computational power and amounts of data perform a different number of epochs, including fractional epochs. A hyperparameter $\lambda$ controls the number of local passes (epochs) the slowest learner in the federation needs to perform before all learners synchronize. For example, $\lambda=2$ has the slowest learner completing two epochs. In a heterogeneous federation with different amounts of data or computational capabilities (e.g., different generation GPUs/CPUs), FedAvg will have fast sites remain idle waiting for the slow sites. In SemiSync there is no idle time. See~\cite{stripelis2022:semisync} for details on the SemiSync FL training policy. 

We do not discuss encryption and security threats (e.g., membership attacks \cite{gupta2021membership}) in this paper, but we refer to our previous work that shows that our FL architecture performs efficiently (only $\sim$7\% overhead) under homomorphic encryption \cite{stripelis2024:patterns} and large federations~\cite{stripelis2023:metisfl}.

\section{Experiments}

We study the performance of centralized and federated training of a deep learning model for binary classification of Alzheimer's disease (AD) status (controls vs AD) from structural T1 MRIs under different amounts of data, and distributions of data across multiple sites. Specifically, we evaluate the effect of having sites with large and small amounts of data, which is typical of scientific consortia. As we will see, such configurations have a significant and surprising effect on the convergence rate and final learning performance (e.g., accuracy) of federated learning. 

\subsection{Data}

We used MRI scans from the first three phases of the Alzheimer’s Disease Neuroimaging Initiative (ADNI) \cite{jack2008alzheimer}. MRIs were preprocessed following the pipeline in Dhinagar et al.\cite{dhinagar2021:sipaim}. We used non-accelerated scans with 9 degrees of freedom. The training, validation, and test set contain 3,293 (1927 controls, 1366 AD), 415 (244 controls, 171 AD), and 454 (262 controls, 192 AD) MRIs, respectively. These sets were subject-disjoint to prevent data leakage. Since ADNI collects MRIs for a subject across multiple visits, the MRIs for a subject only appear in one of the train, validation, or test datasets. All splits in centralized and federated experiments were stratified on diagnosis. The experiments were replicated across ten runs with different seeds.

\subsection{Deep learning Model}

\begin{figure}[tbp]
\centering
\includegraphics[width=\columnwidth]{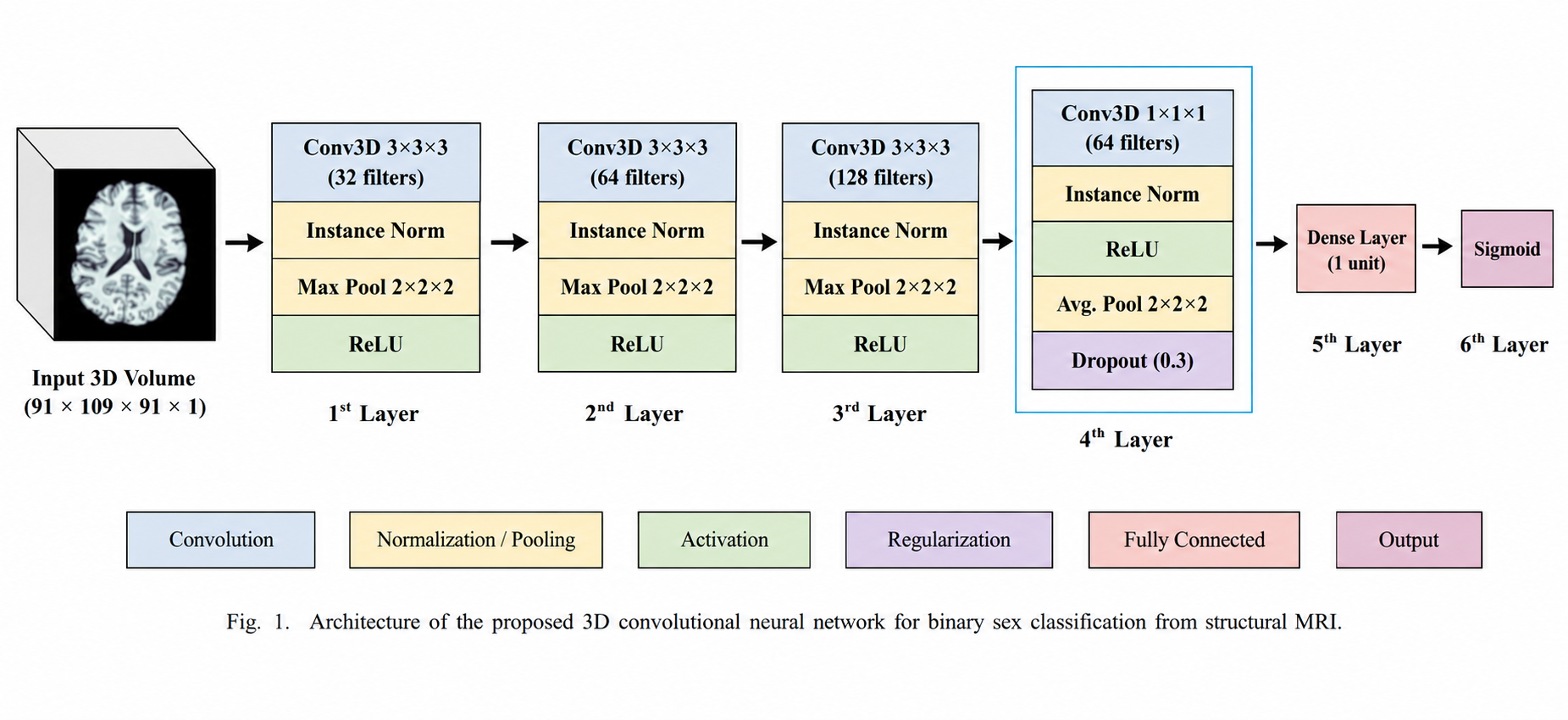}
\caption{3D Convolutional Neural Network for binary classification of AD status from T1 MRI.}
\label{fig:neural-architecture}
\end{figure}

We use a 3D convolutional neural network for binary classification of volumetric neuroimaging data (Figure~\ref{fig:neural-architecture}). The network accepts a preprocessed input volume of size 91 x 109 x 91 × 1 and comprises three sequential 3D convolutional blocks, each using a 3 × 3 × 3 kernel followed by instance normalization, 2 x 2 x 2 max pooling, and ReLU activation. The number of filters increases progressively from 32 to 64 and 128 to learn hierarchical spatial features. After the feature-extraction blocks, a 1 x 1 x 1 convolution with 64 filters is applied for channel compression, followed by instance normalization, ReLU activation, and 2 x 2 x 2 average pooling. The resulting representation is regularized using dropout (0.3), flattened, and passed to a final sigmoid-activated dense layer to produce the binary output. The network is trained with binary cross-entropy loss using the Adam optimizer with a learning rate of 5x$10^{-5}$. Performance is monitored using accuracy and AUC. Early stopping and checkpointing are used to improve training stability and preserve the best validation model.

\subsection{Centralized Experiments}

\begin{figure}[tbp]
\centering
\includegraphics[width=\columnwidth]{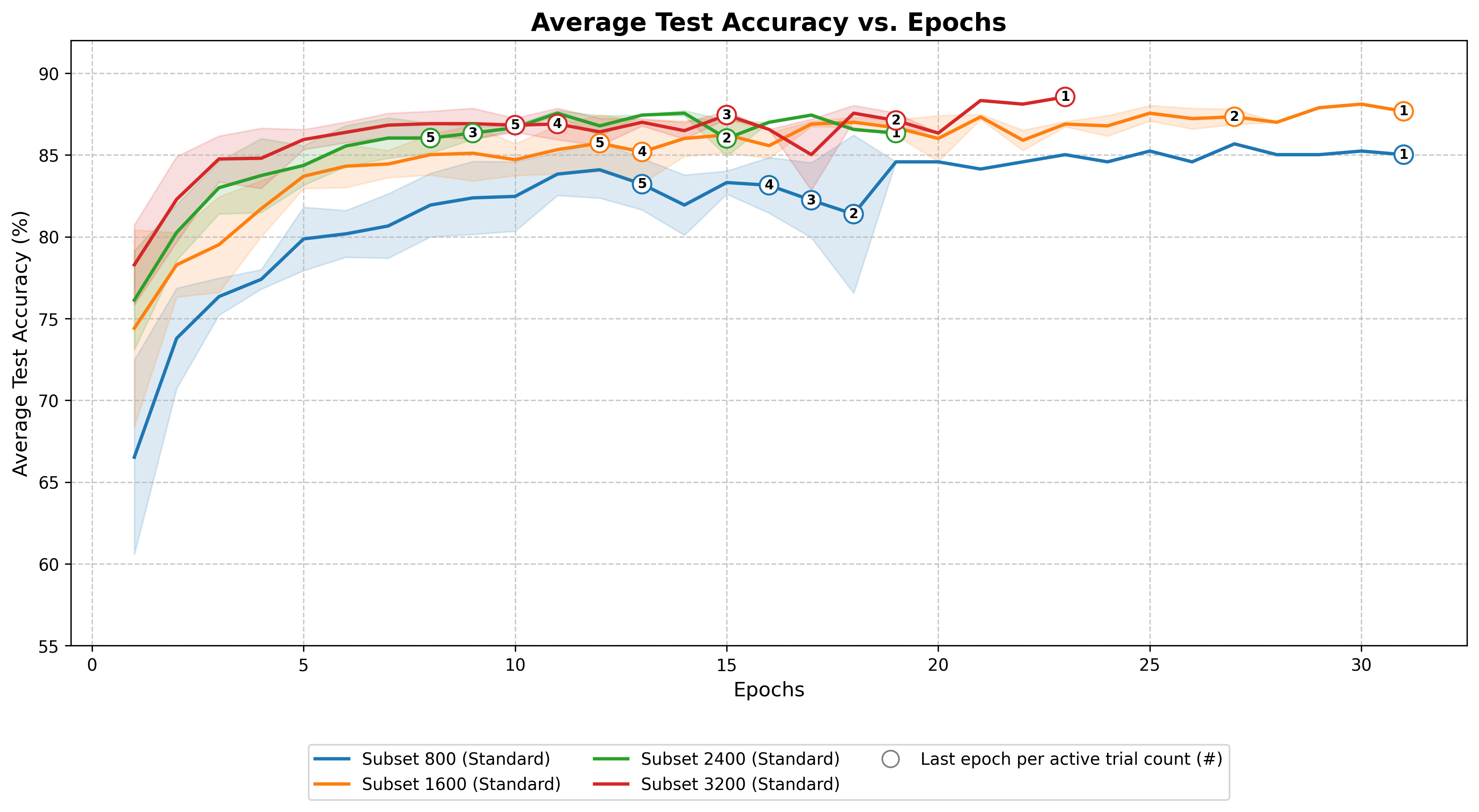}
\caption{Centralized performance (accuracy) of AD classification for different amounts of data (average over 5 runs, shaded band of 1 standard deviation).}
\label{fig:centralized}
\end{figure}

To establish a baseline, we performed centralized experiments with different amounts of data: 600, 700, 800, 1000, 1600, 2400, and 3200 MRIs from ADNI. Figure~\ref{fig:centralized} shows the convergence rates (training epochs vs accuracy) for 5 runs of different training set sizes. In this plot, we used early stopping (patience=5), so each run terminated after a different number of epochs. The circled numbers in the figure indicate how many runs are averaged up to that point. Unsurprisingly, larger training sets result in a greater final accuracy, which plateaus $\sim$86\% for the larger datasets.   

\subsection{Federated Learning Experiments with Homogeneous and Heterogeneous Data Amounts}

We set up a federation of 8 sites and study the effect on federated learning of different data amounts across multiple sites. Specifically,  we compare the convergence rate and final accuracy for homogeneous data distributions (same data amount per site) vs heterogeneous data distributions (one large site and several small sites) for different total amounts of examples across the federation. We consider 600, 700, 800, 1000, 1600, 2400, and 3200 total examples across the sites of the federation, In the homogeneous case the data are distributed uniformly, namely, 8x100 (i.e., 8 sites with 100 MRIs each), 8x200, 8x300, and 8x400. In the heterogeneous case the data distributions are: 7x50+1×450 (i.e., 7 sites with 50 MRIs each, plus one large site with 450 MRIs), 7×150+1x550, 7x250+1x650, and 7×350+1x750.
We also considered federations with lower amounts of data, specifically with 600, 700, and 1000 total MRIs: for the homogeneous case, we consider 6 sites of 100 MRIs, 7 sites of 100 MRIS, and 5 sites of 200 MRIs; and for the  heterogeneous case, sites with 3x50+1x450, 5x50+1x450, and 3x150+ 550 (i.e., 3 sites with 150 MRIs each, and 1 site with 550 MRIs).  

We performed experiments with homogeneous and heterogeneous data distributions and with the FedAvg and SemiSync training policies. SemiSync used $\lambda=1$, that is, the synchronization period is the time it takes for the slowest learner to complete one epoch on its local training set. Since the GPUs of all sites were identical, this means that the largest site is the slowest and sets the synchronization period. 

\begin{figure}[htbp]
    \centering

    \begin{subfigure}[t]{0.48\columnwidth}
        \centering
        \includegraphics[width=\linewidth]{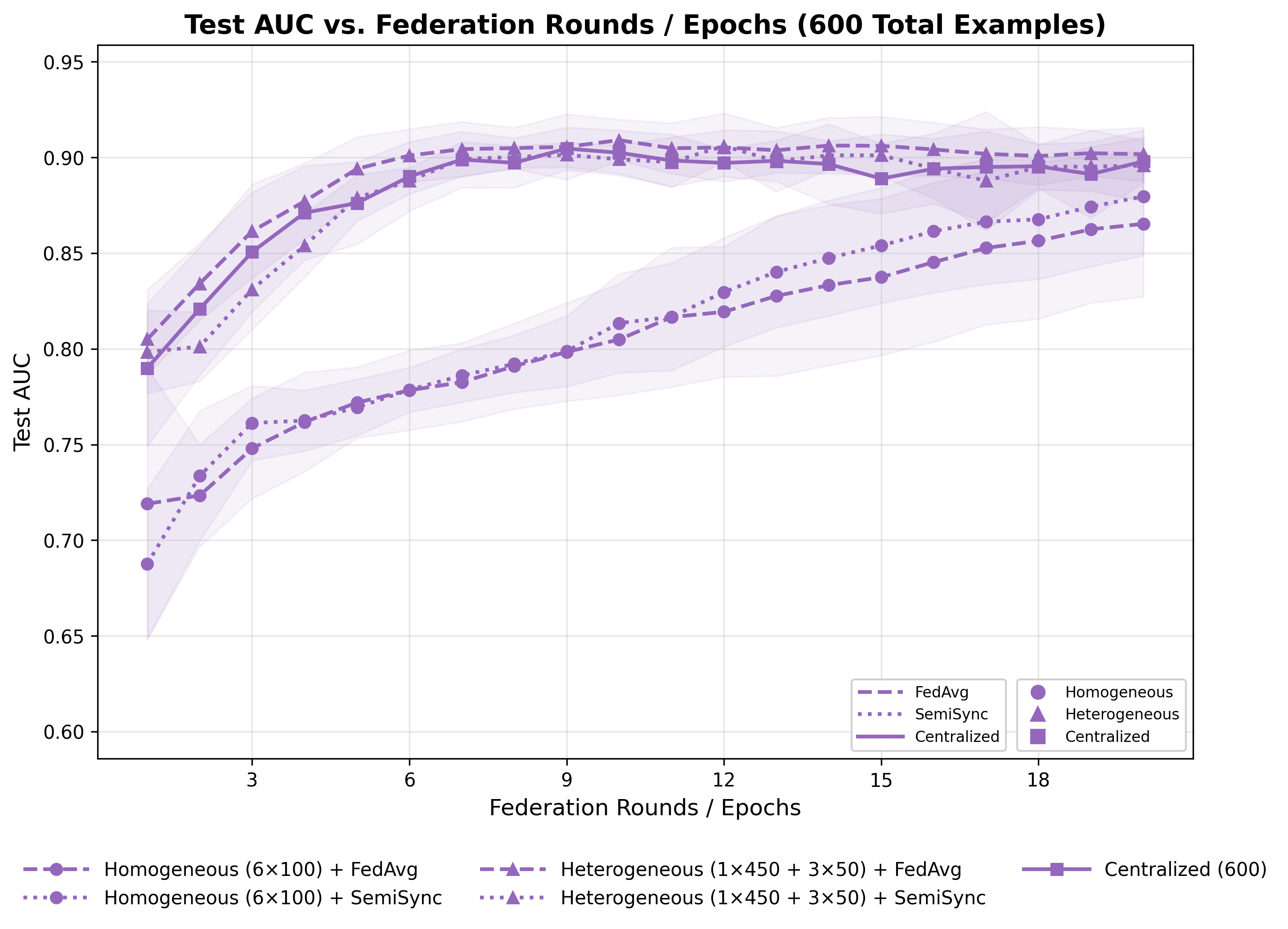}
        \caption{600 total examples}
    \end{subfigure}
    \hfill
    \begin{subfigure}[t]{0.48\columnwidth}
        \centering
        \includegraphics[width=\linewidth]{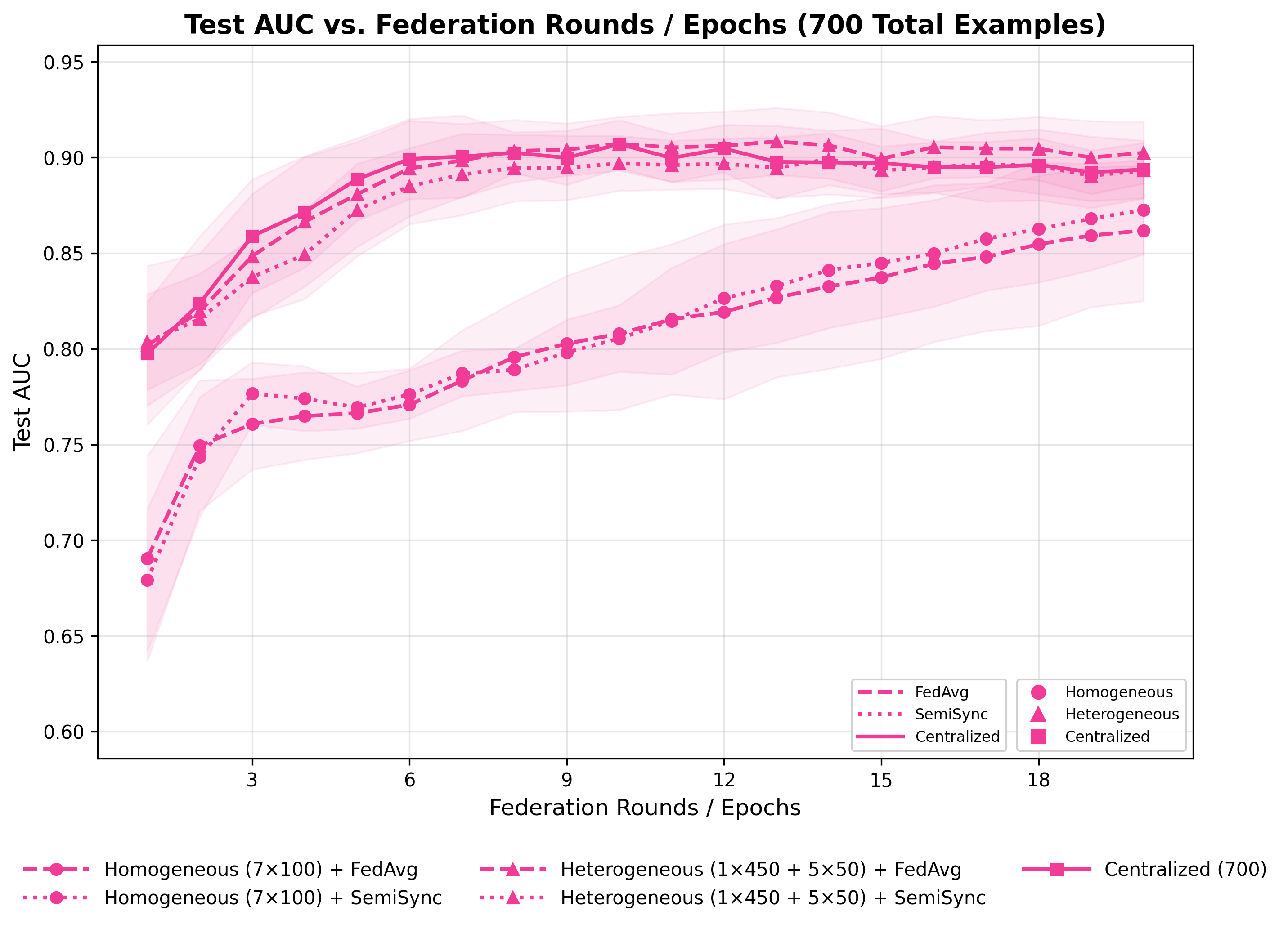}
        \caption{700 total examples}
    \end{subfigure}

    \medskip

    \begin{subfigure}[t]{0.48\columnwidth}
        \centering
        \includegraphics[width=\linewidth]{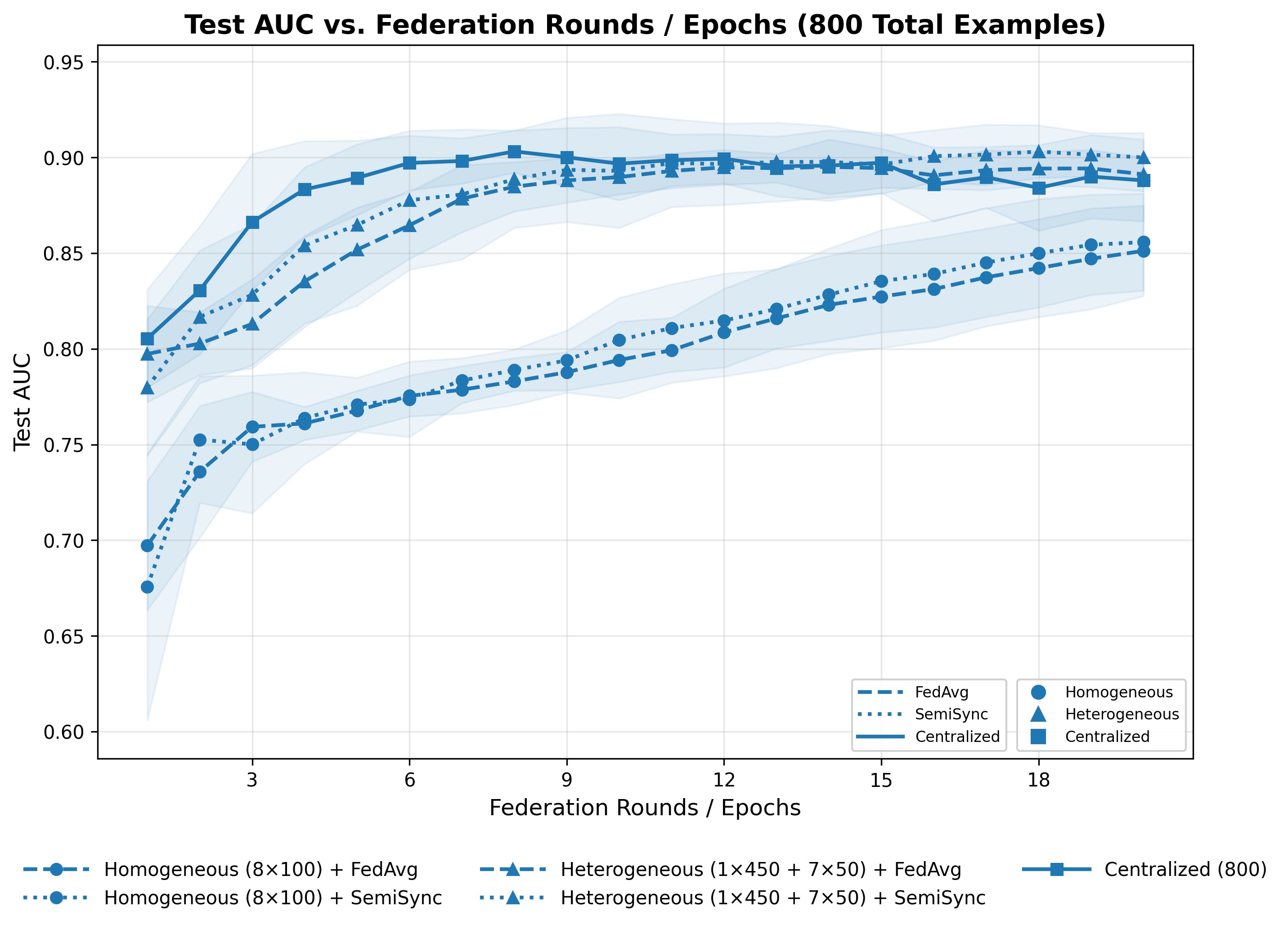}
        \caption{800 total examples}
    \end{subfigure}
    \hfill
    \begin{subfigure}[t]{0.48\columnwidth}
        \centering
        \includegraphics[width=\linewidth]{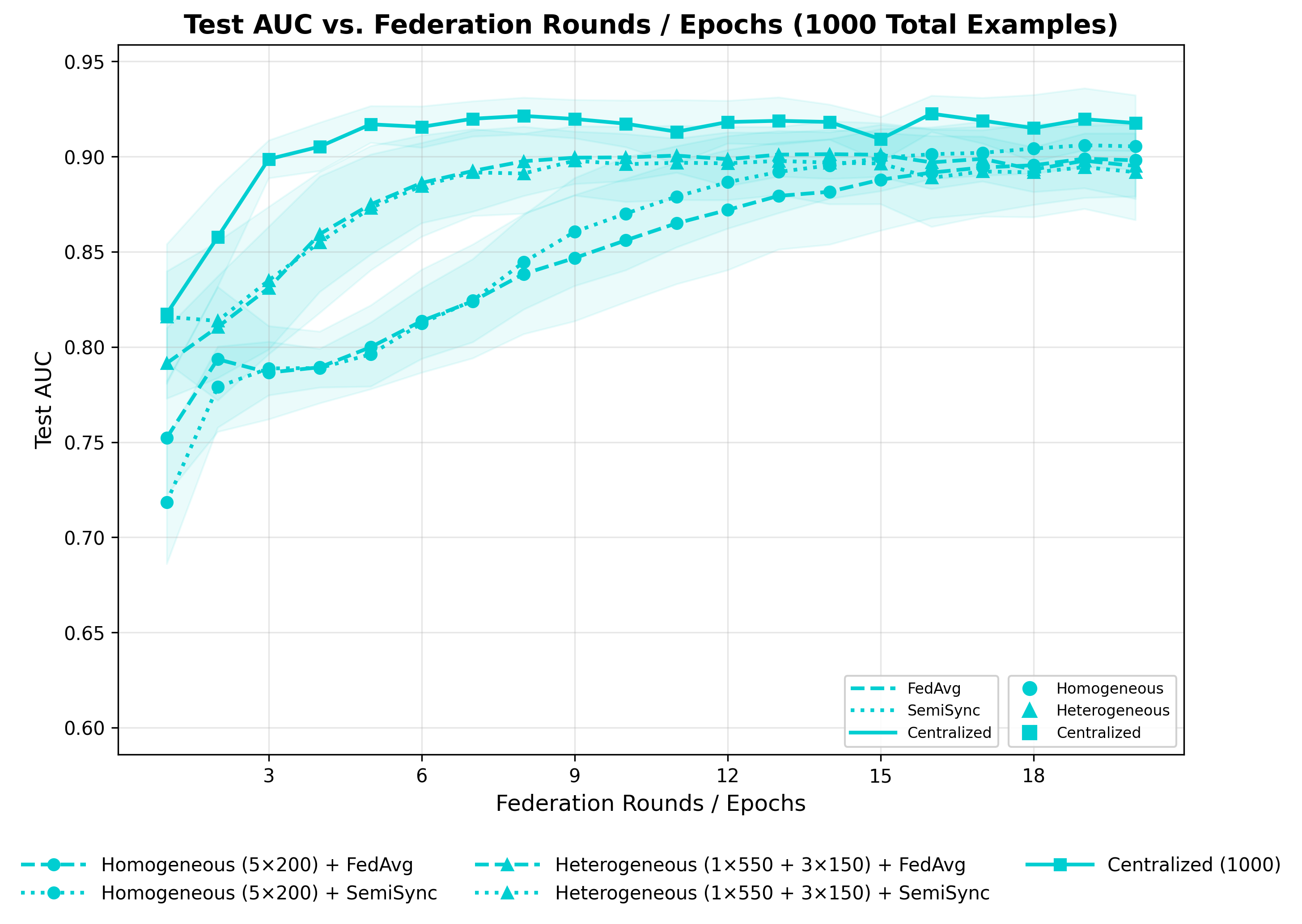}
        \caption{1000 total examples}
    \end{subfigure}

    \medskip

    \begin{subfigure}[t]{0.48\columnwidth}
        \centering
        \includegraphics[width=\linewidth]{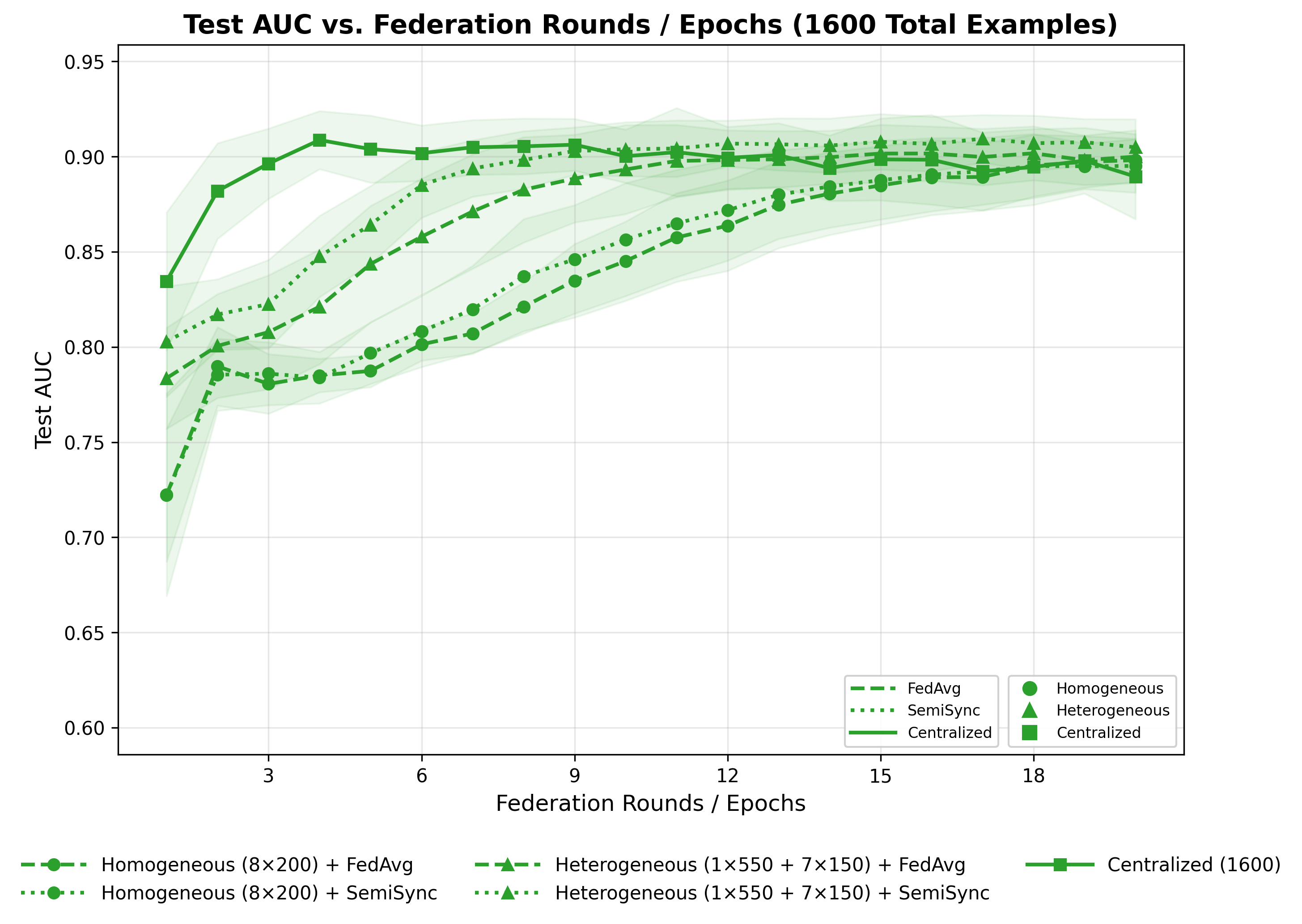}
        \caption{1600 total examples}
    \end{subfigure}
    \hfill
    \begin{subfigure}[t]{0.48\columnwidth}
        \centering
        \includegraphics[width=\linewidth]{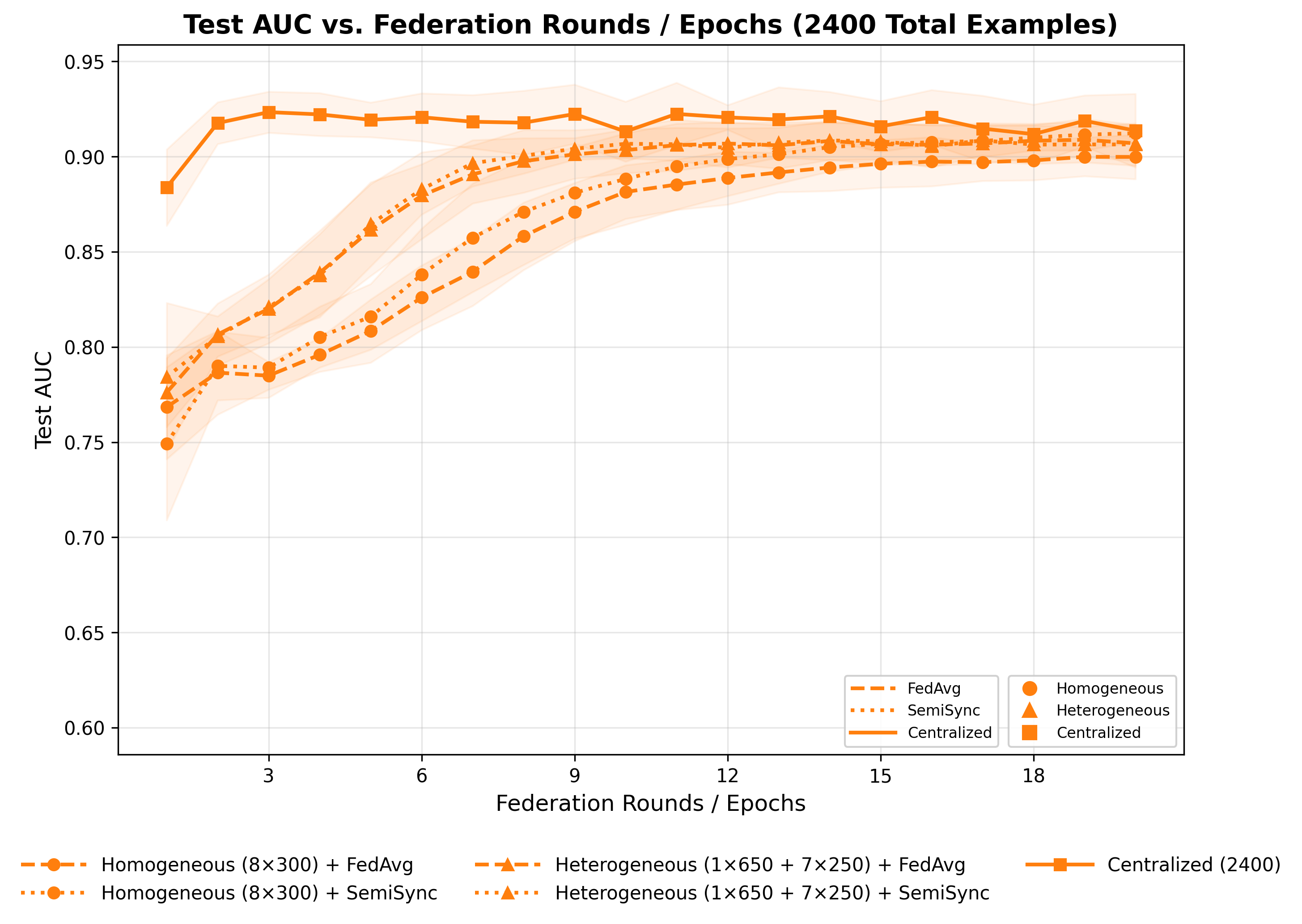}
        \caption{2400 total examples}
    \end{subfigure}

    \medskip

    \begin{subfigure}[t]{0.48\columnwidth}
        \centering
        \includegraphics[width=\linewidth]{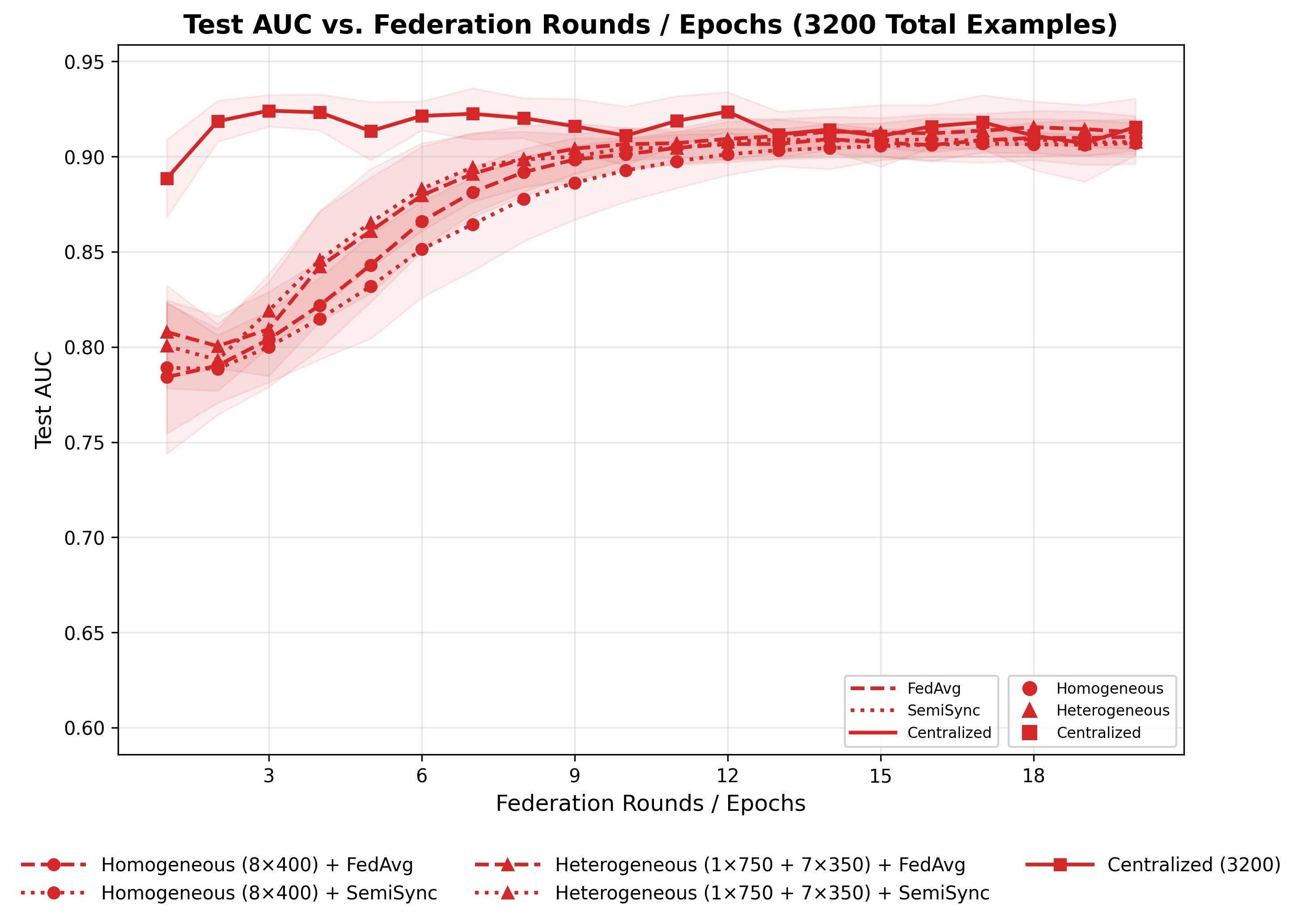}
        \caption{3200 total examples}
    \end{subfigure}
    
    \medskip
    \begin{subfigure}[t]{\columnwidth}
        \centering
        \includegraphics[width=\linewidth]{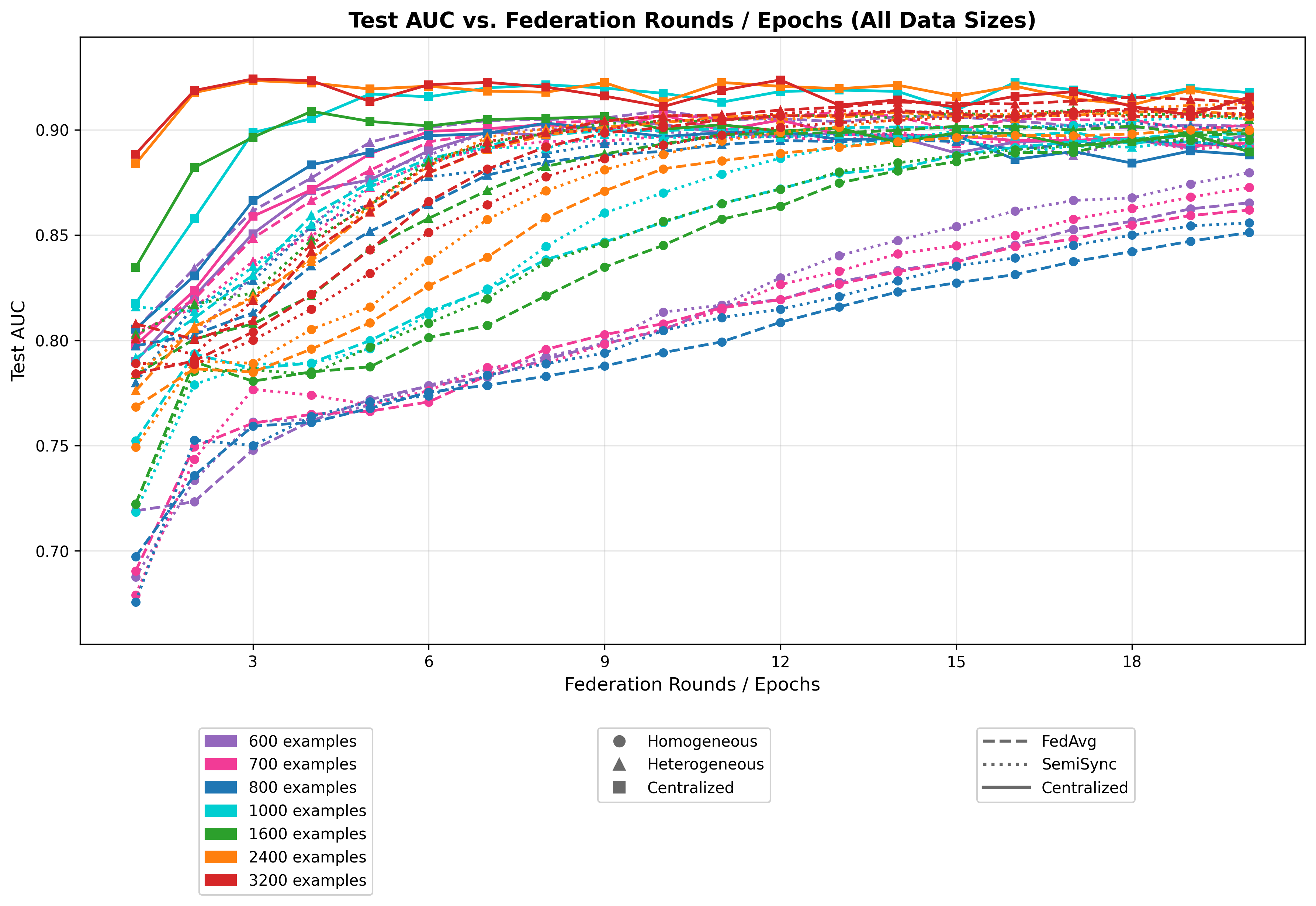}
        \caption{Combined}
    \end{subfigure}

    \caption{Federated Learning performance (AUC) of AD classification for different amounts of data for centralized, homogeneous, and heterogeneous environments (average over 10 runs, shaded band of 1 standard deviation).}
    \label{fig:federated-8panel-auc}
\end{figure}

\begin{figure}[htbp]
    \centering

    \begin{subfigure}[t]{0.48\columnwidth}
        \centering
        \includegraphics[width=\linewidth]{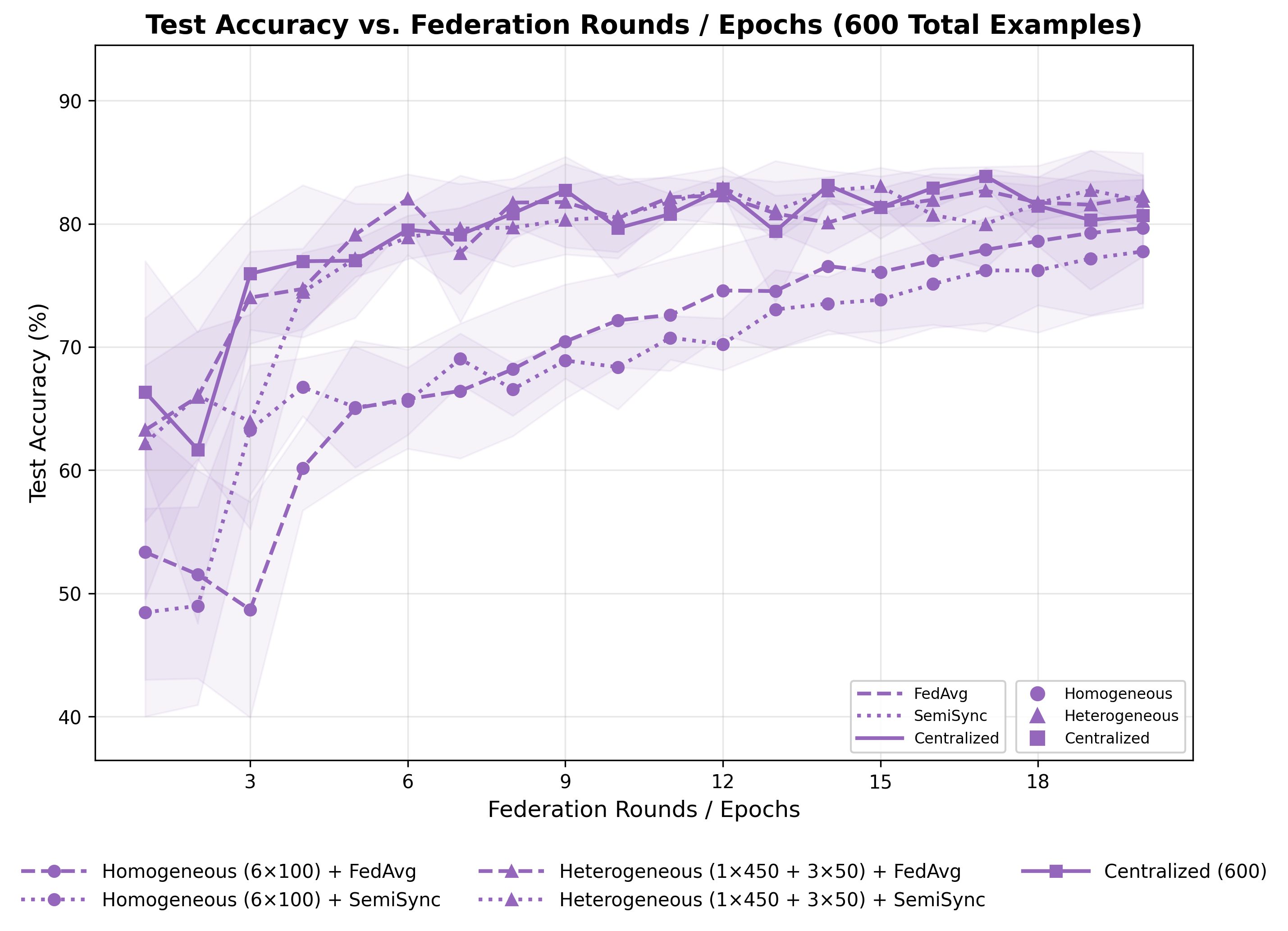}
        \caption{600 total examples}
    \end{subfigure}
    \hfill
    \begin{subfigure}[t]{0.48\columnwidth}
        \centering
        \includegraphics[width=\linewidth]{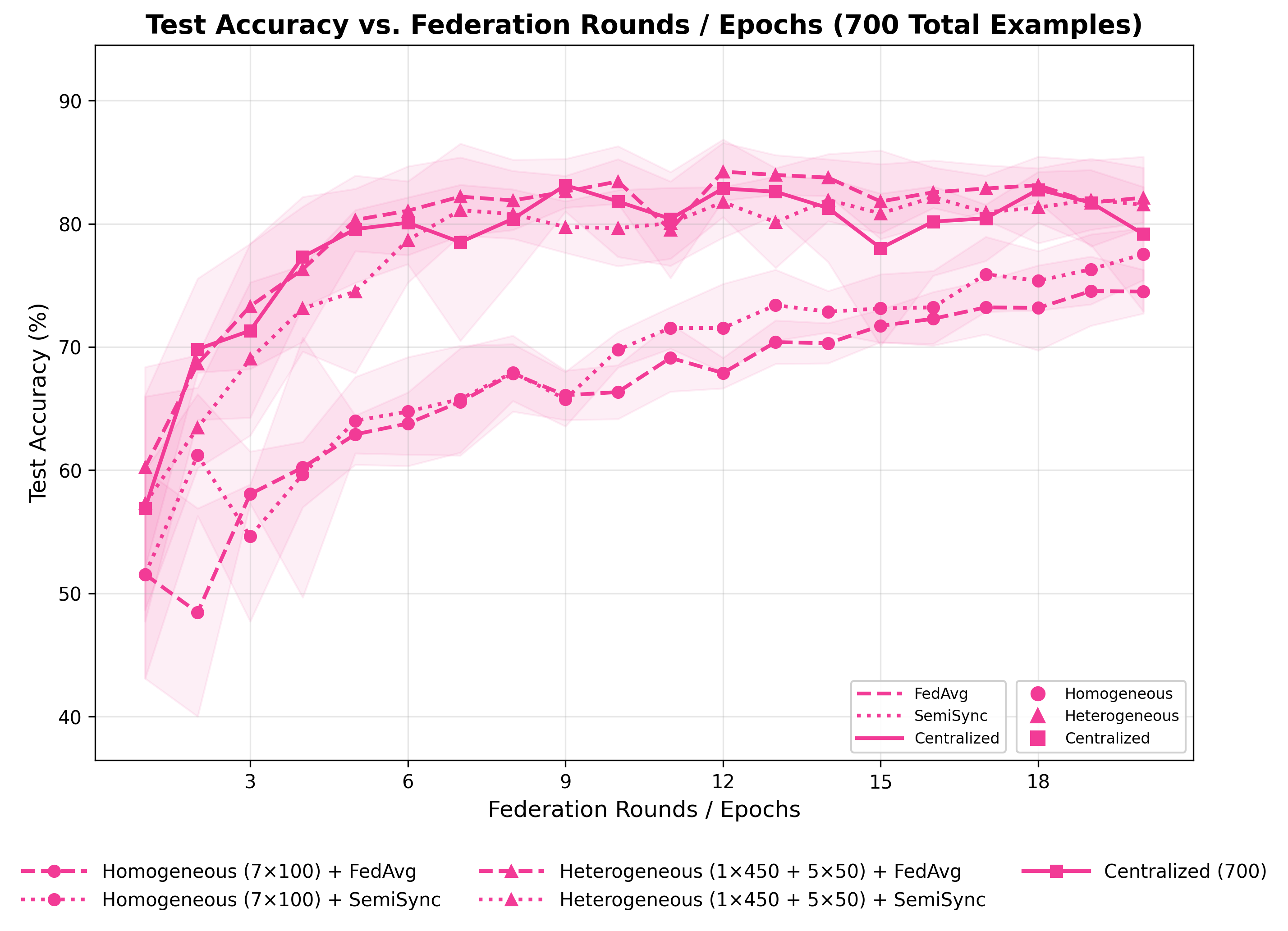}
        \caption{700 total examples}
    \end{subfigure}

    \medskip

    \begin{subfigure}[t]{0.48\columnwidth}
        \centering
        \includegraphics[width=\linewidth]{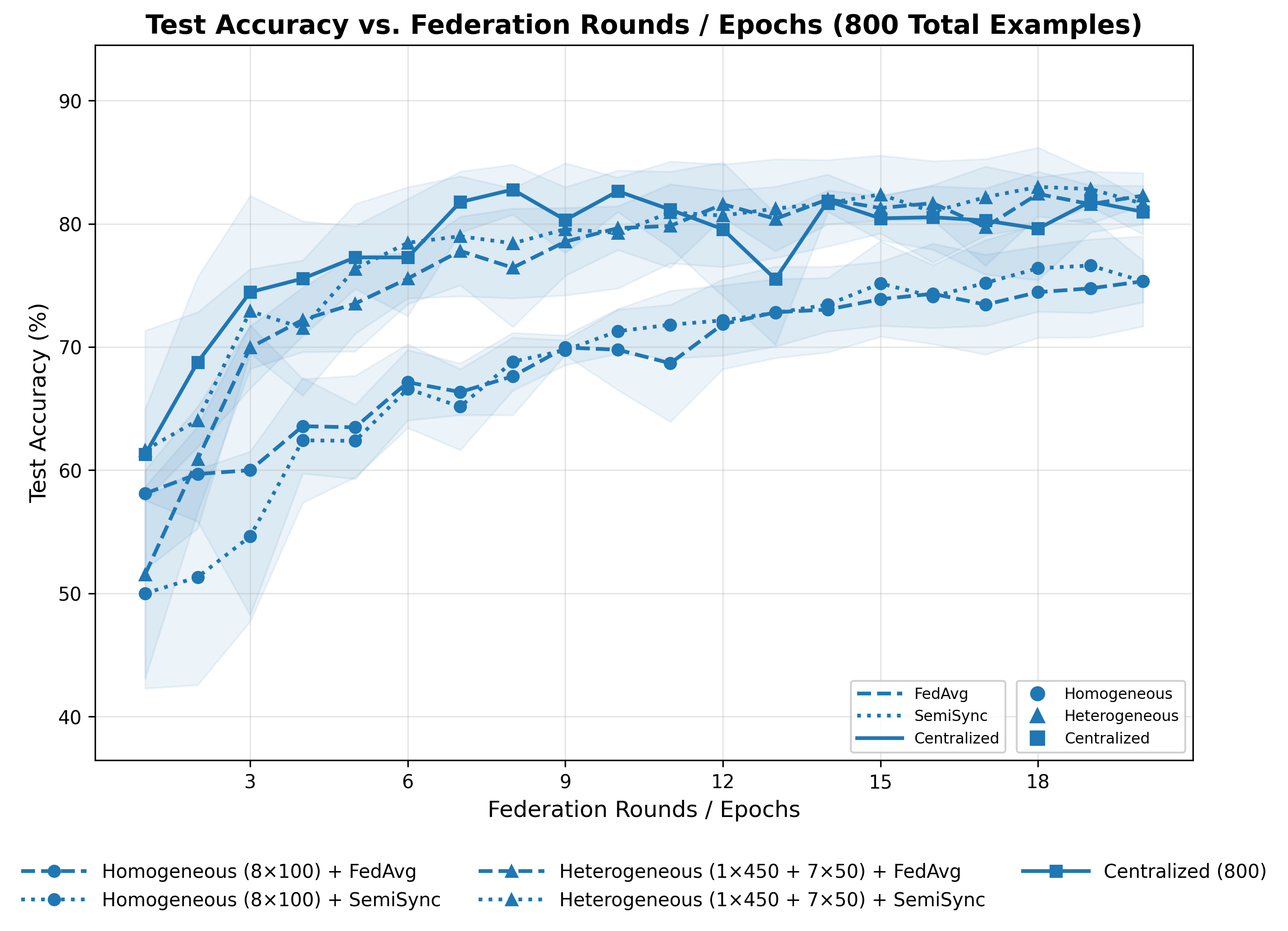}
        \caption{800 total examples}
    \end{subfigure}
    \hfill
    \begin{subfigure}[t]{0.48\columnwidth}
        \centering
        \includegraphics[width=\linewidth]{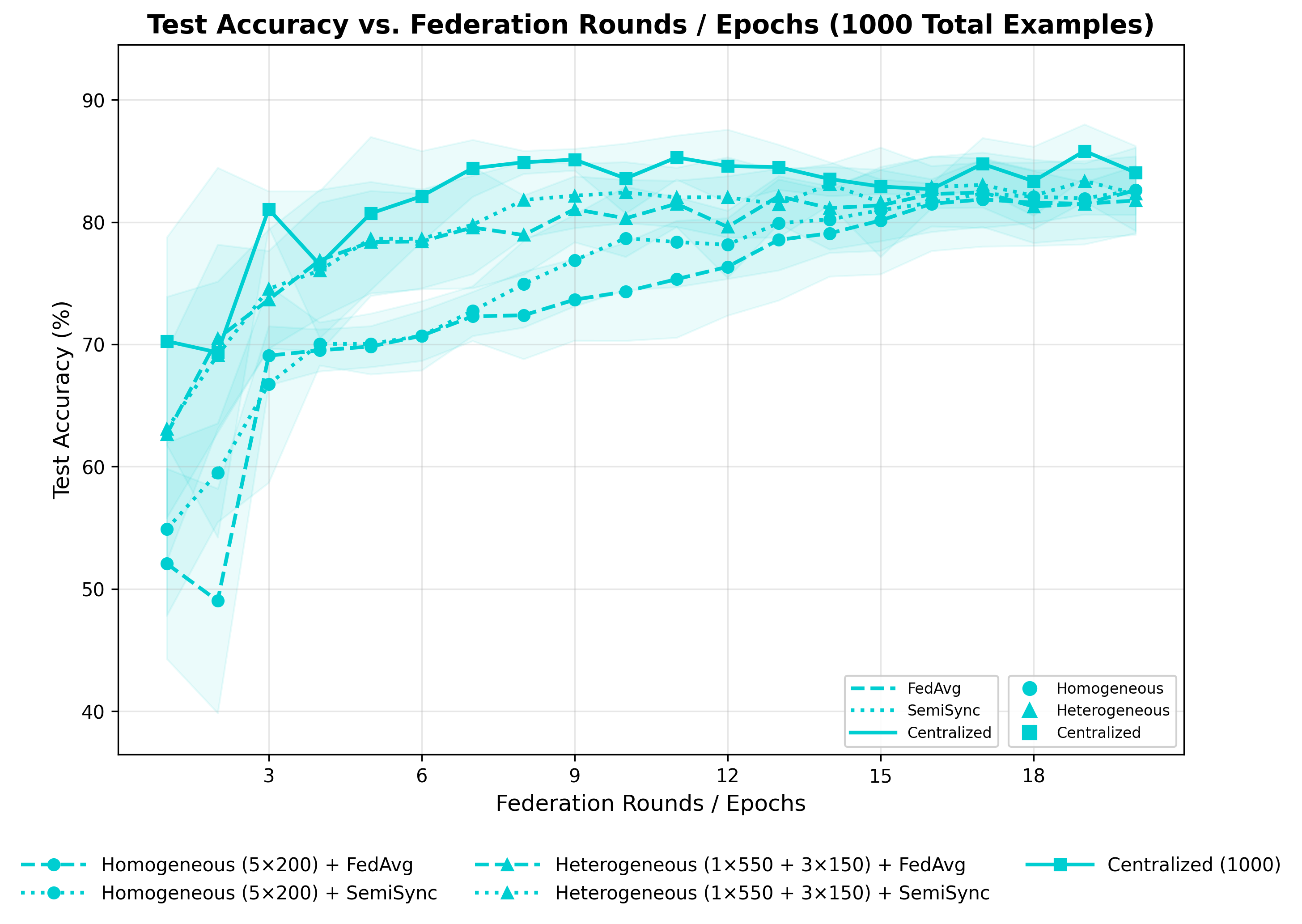}
        \caption{1000 total examples}
    \end{subfigure}

    \medskip

    \begin{subfigure}[t]{0.48\columnwidth}
        \centering
        \includegraphics[width=\linewidth]{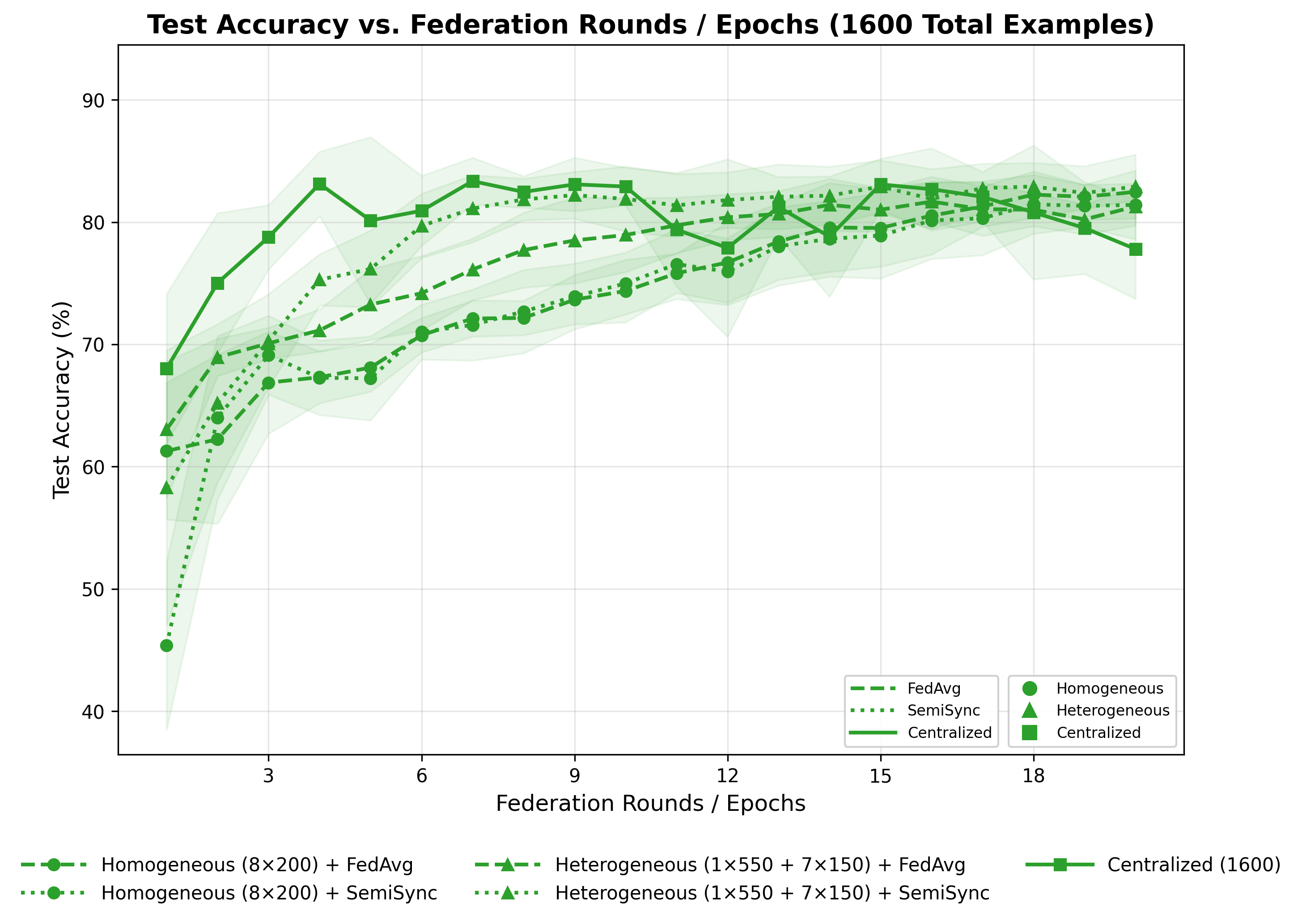}
        \caption{1600 total examples}
    \end{subfigure}
    \hfill
    \begin{subfigure}[t]{0.48\columnwidth}
        \centering
        \includegraphics[width=\linewidth]{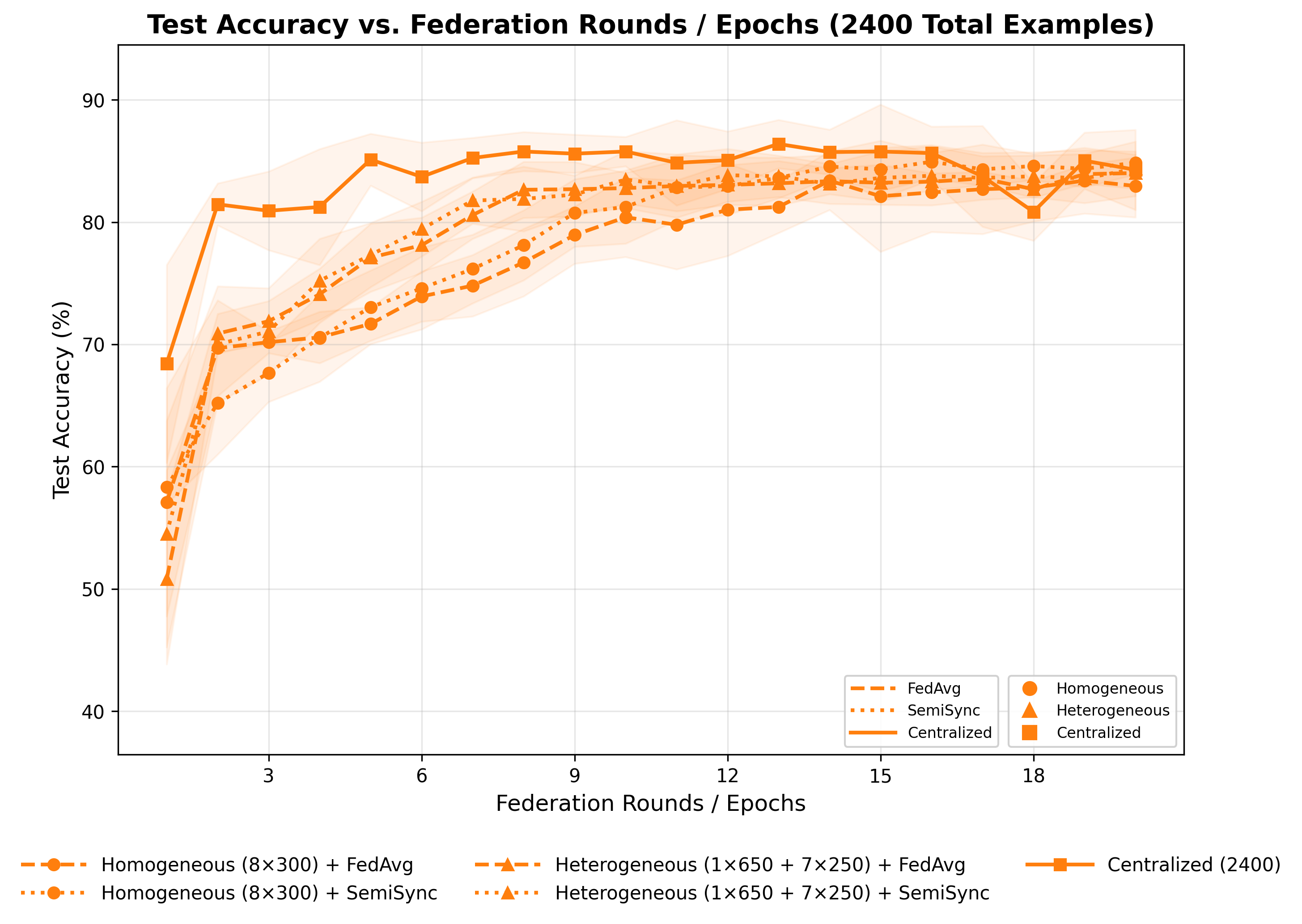}
        \caption{2400 total examples}
    \end{subfigure}

    \medskip

    \begin{subfigure}[t]{0.48\columnwidth}
        \centering
        \includegraphics[width=\linewidth]{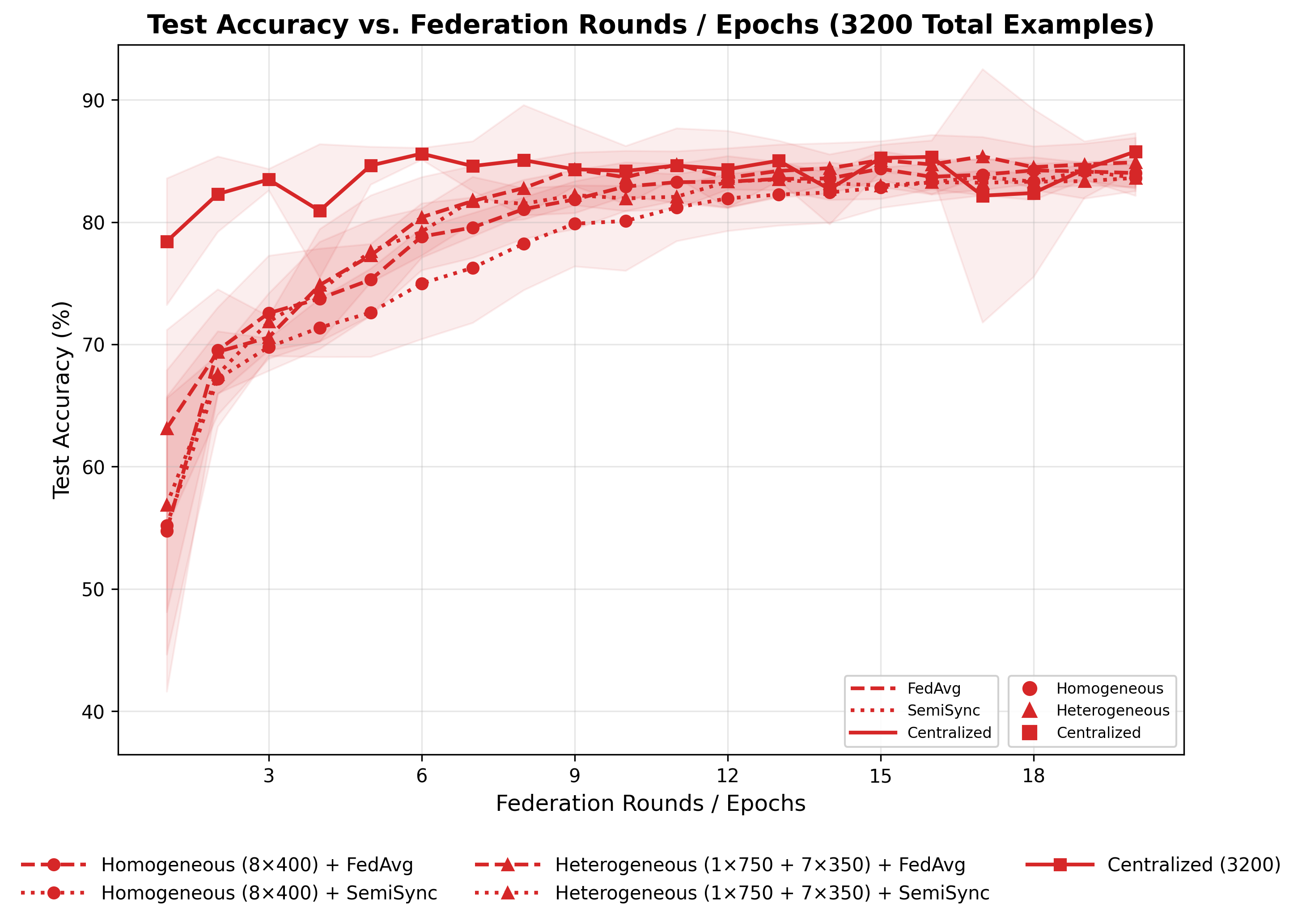}
        \caption{3200 total examples}
    \end{subfigure}

    \medskip
    \begin{subfigure}[t]{\columnwidth}
        \centering
        \includegraphics[width=\linewidth]{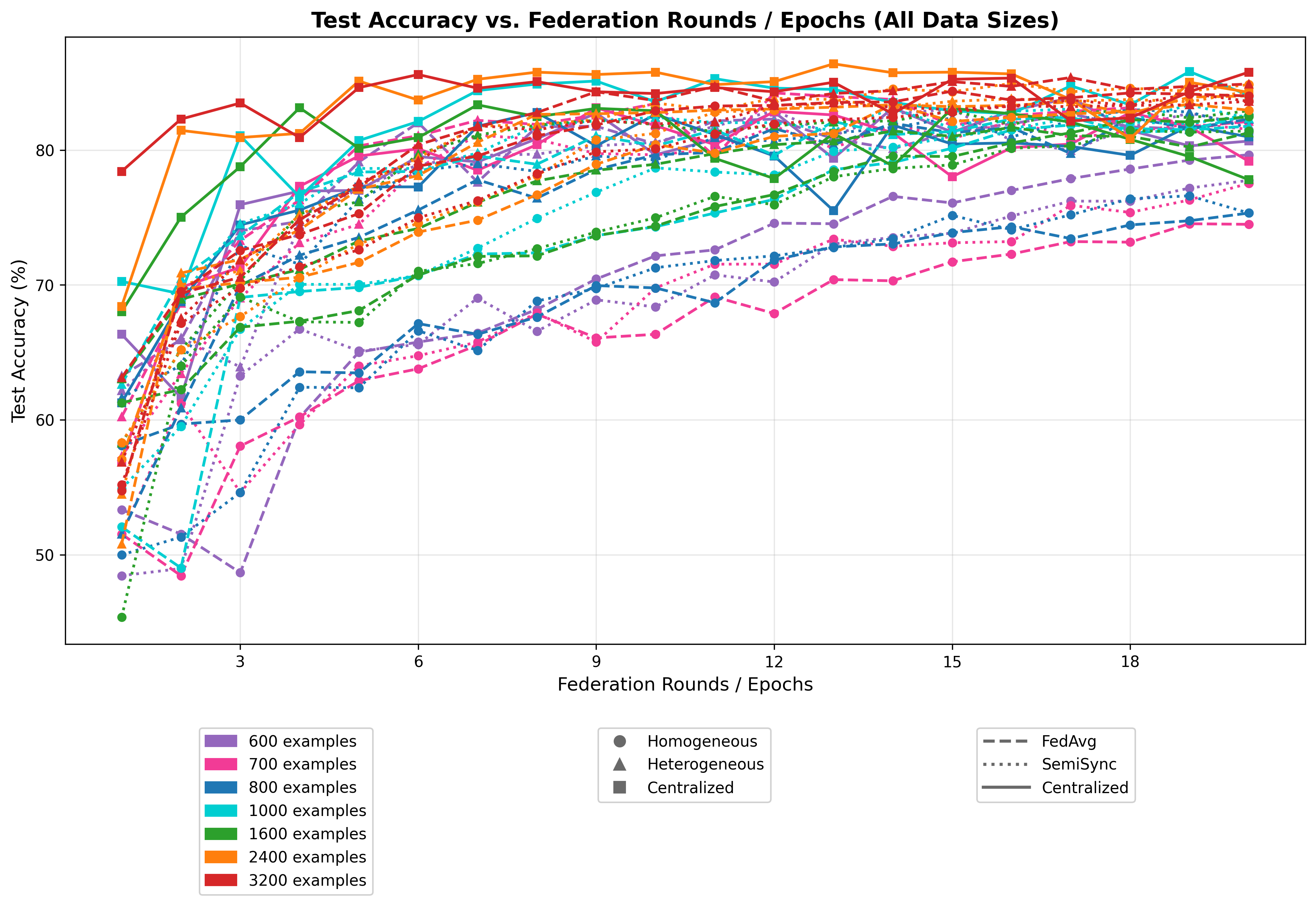}
        \caption{Combined}
    \end{subfigure}

    \caption{Federated Learning performance (Accuracy) of AD classification for different amounts of data for centralized, homogeneous and heterogeneous environments (average over 10 runs, shaded band of 1 standard deviation).}
    \label{fig:federated-8panel-accuracy}
\end{figure}

Figures~\ref{fig:federated-8panel-auc} and~\ref{fig:federated-8panel-accuracy} show the convergence rates for learning performance for the Area Under the ROC Curve (AUC) and accuracy, respectively, on the held-out test set across different total amounts of data (MRIs) and distributions. Each line is the average of 10 runs with a shaded band of 1 standard deviation. Surprisingly, the heterogeneous case with one large site and the rest small sites performs better than the homogeneous case (all sites with the same amount of data). This is more pronounced in low data regimes. For example, consider the case of 800 total MRIs, the federation with one large site (450 MRIs) and 7 very small sites (50 MRIs each) outperforms the homogeneous case (8x100) by $\sim$4 points of AUC (~86 vs 90). When sites have more data the final performance tends to converge.

\begin{figure}[htbp]
\centering
    \begin{subfigure}[t]{\columnwidth}
        \centering
        \includegraphics[width=\linewidth]{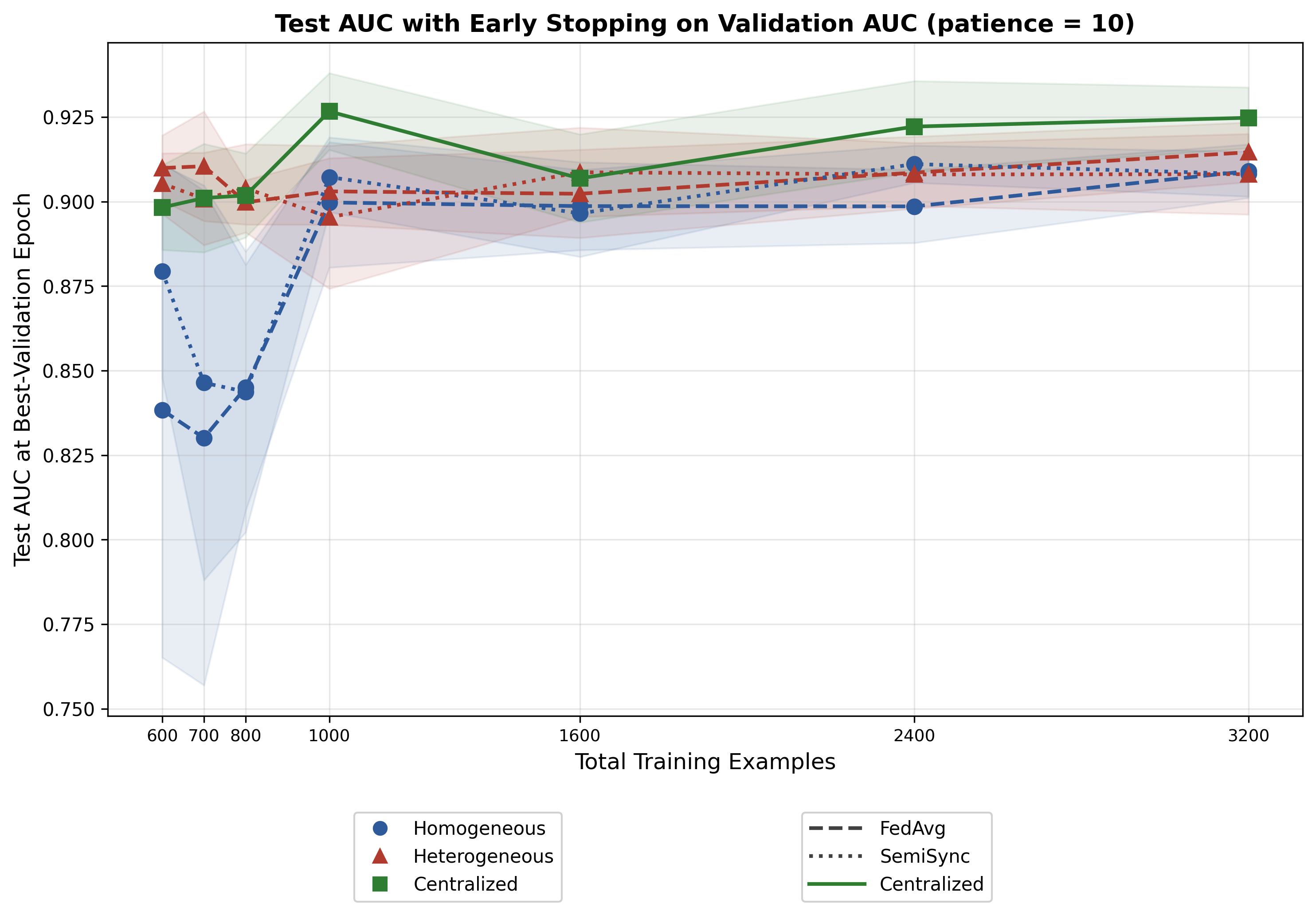}
        \caption{AUC (ROC)}
    \end{subfigure}
    
    \medskip
    \begin{subfigure}[t]{\columnwidth}
        \centering
        \includegraphics[width=\linewidth]{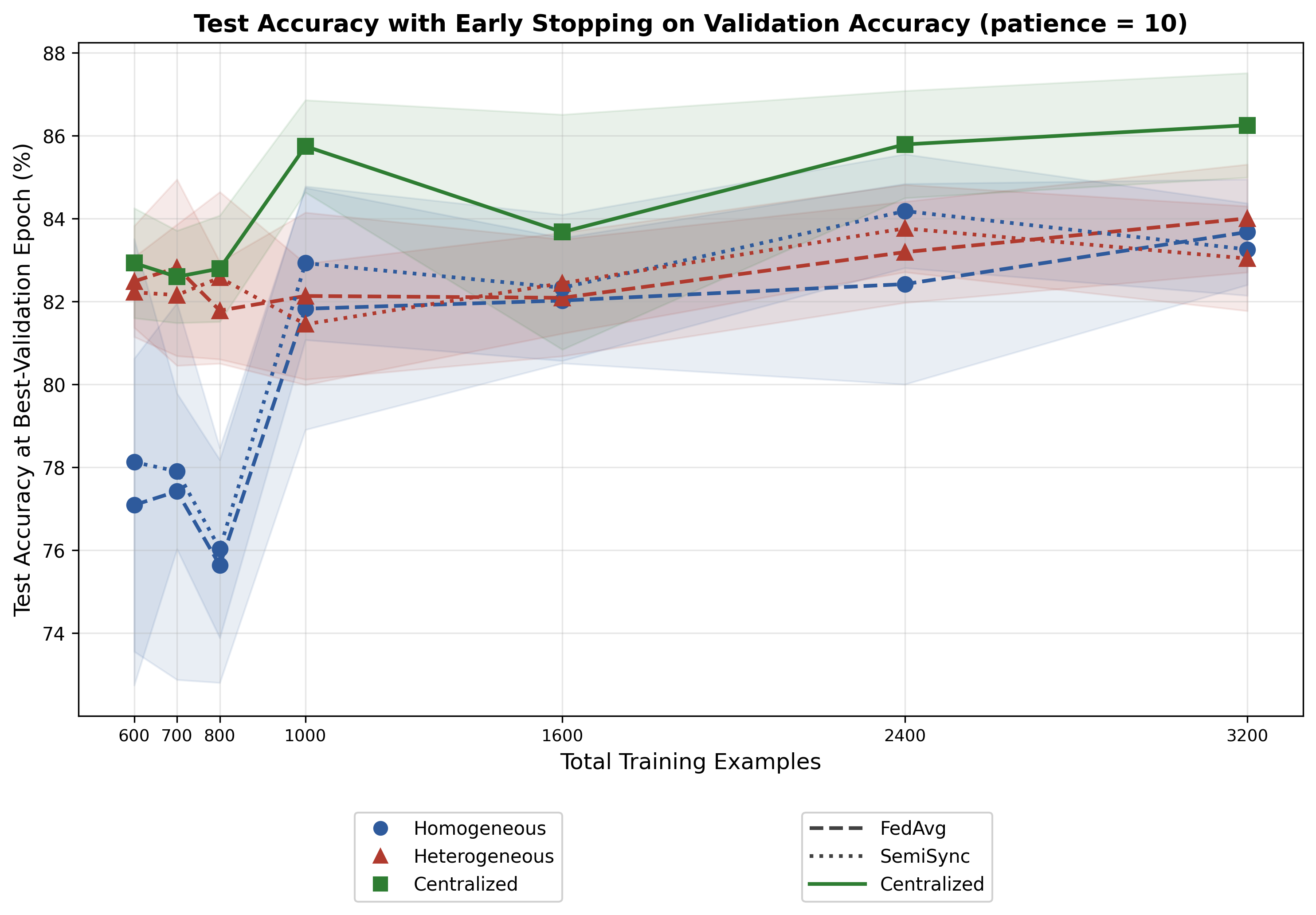}
        \caption{Accuracy}
    \end{subfigure}
\caption{Federated Learning performance, Accuracy and AUC (ROC), of AD classification for different amounts of data for centralized, homogeneous and heterogeneous environments. Average test accuracy over 5 runs using early stopping, shaded band of 1 standard deviation.}
\label{fig:scaling-early}
\end{figure}

Figure~\ref{fig:scaling-early} shows learning performance, AUC ROC (a) and Accuracy (b), on the held-out test set for centralized, homogeneous, and heterogeneous environments as we increase the total amount of data (MRIs) under early stopping (patience = 10, average of 10 runs, 1 standard deviation shaded area). Heterogeneous federations outperform homogeneous, particularly in low data regimes, and reach performance close to centralized training.

\subsection{Discussion}

We have studied the effect of different data amounts across sites in federated learning. Specifically, heterogeneous federations with one large site in the federation with the remaining sites having small amounts of data, compared to homogeneous federations where each site has the same amount of data, while keeping the total amount of data across the federations constant.
Surprisingly, having a large site in the federation facilitates exploiting sites with small amounts of data yielding better learning performance than homogeneous federations,  particularly in cases with smaller amounts of total data. 
We hypothesize that the reason for this phenomenon is that the model learned on the large site is of higher quality and acts as a form of online model transfer to the models learned over the small sites as the federated aggregation progresses. This behavior bodes well for the incorporation of small sites in large consortia. 

In future work, we plan to study more extensively the value of data across sites in federated consortia, analyzing both the amounts of data in different sites as well as data diversity. We plan to evaluate the impact of pretraining on a related task, e.g., sex classification, on the performance of the federation. Pretraining may allow to successfully exploit data in sites with even smaller amounts of data.

\section*{Acknowledgment}

This work was supported in part by NIH grants RF1AG081571, U01AG068057, R01MH131806, and S10OD032285.



%
%
%

\bibliographystyle{IEEEtran}
\bibliography{refs}

\end{document}